\documentclass[draftclsnofoot,onecolumn]{IEEEtran}
\usepackage{amsmath,amsfonts}
\usepackage{algorithmic}
\usepackage{algorithm}
\usepackage{array}
\usepackage[caption=false,font=normalsize,labelfont=sf,textfont=sf]{subfig}
\usepackage{textcomp}
\usepackage{stfloats}
\usepackage{url}
\usepackage{verbatim}
\usepackage{graphicx}

\usepackage{cite}
\usepackage{booktabs}
\usepackage{multirow}
\usepackage{multicol}
\usepackage[table]{xcolor}

\begin{document}

\title{Rectified Iterative Disparity for Stereo Matching}

\author{Weiqing Xiao, Wei Zhao$^{\ast}$
\thanks{Weiqing Xiao is with the School of Electronic Information Engineering, 
Beihang University, Beijing 100191, China, (e-mail: xiaowqtx@buaa.edu.cn)}
\thanks{\textit{Corresponding author: Wei Zhao.}}
}

\markboth{Rectified Iterative Disparity for Stereo Matching}%
{Shell \MakeLowercase{\textit{et al.}}: A Sample Article Using IEEEtran.cls for IEEE Journals}

\IEEEpubid{0000--0000/00\$00.00~\copyright~2021 IEEE}

\maketitle

\section{recommendation}
  The updated version will be released soon!!!

\begin{abstract}
  Both uncertainty-assisted and iteration-based methods have 
  achieved great success in stereo matching. However, existing 
  uncertainty estimation methods take a single image and the 
  corresponding disparity as input, which imposes higher demands on the 
  estimation network. In this paper, we propose \emph{Cost volume-based disparity Uncertainty Estimation} (UEC). Based on the rich similarity 
  information in the cost volume coming from the image pairs, the proposed 
  UEC can achieve competitive performance with low computational cost. 
  Secondly, we propose two methods of uncertainty-assisted disparity estimation, 
  \emph{Uncertainty-based Disparity Rectification} (UDR) and \emph{Uncertainty-based 
  Disparity update Conditioning} (UDC). These two methods optimise the 
  disparity update process of the iterative-based approach without 
  adding extra parameters. In addition, we propose \emph{Disparity Rectification loss}
  that significantly improves the accuracy of small amount of 
  disparity updates. We present a high-performance stereo architecture, 
  \emph{DR Stereo}, which is a combination of the proposed methods. Experimental results from 
  SceneFlow, KITTI, Middlebury 2014, and ETH3D show that DR-Stereo 
  achieves very competitive disparity estimation performance.
\end{abstract}

\begin{IEEEkeywords}
  3D computer vision, Stereo, Uncertainty, Iteration 
\end{IEEEkeywords}

\section{Introduction}
\label{sec:intro}

Depth perception is the basis for computer vision and graphics research 
in 3D scenes. High-precision depth information is vital for fields such 
as 3D reconstruction, autonomous driving, and robotics. Stereo matching is 
an efficient and low-cost depth estimation method that aims at estimating 
the pixel horizontal displacement map, also known as the disparity 
map, between the corrected left and right image pairs. 
Given the camera calibration parameters, we can calculate the depth map from the disparity. 
In recent years, many learning-based stereo networks~\cite{chang2018pyramid,xu2021bilateral,li2021revisiting,duggal2019deeppruner,gu2020cascade,liang2019stereo} have 
achieved encouraging success in terms of quality and efficiency of 
disparity estimation. 
In general, the stereo matching algorithm consists of four steps: 
matching feature extraction, matching cost computation, cost aggregation 
and disparity optimization.

Current research in learning-based stereo networks focuses on the quality 
and efficiency of disparity estimation. 3D convolution-based 
methods~\cite{chang2018pyramid,guo2019group,xu2022attention} use 3D convolution to aggregate and 
regularize 4D cost volume, and then regress the disparity map 
from the regularized cost volume. These methods effectively 
encode context information as well as stereo geometry information 
and achieves good performance. However, the cost aggregation and 
regularization require a large number of 3D convolutions, 
which limits its practical applicability. Correlation volume-based 
methods~\cite{lipson2021raft,xu2023iterative,mayer2016large,yang2018segstereo} use 2D convolution 
instead of 3D for cost aggregation, which saves computational cost but also reduces accuracy. 
Iterative based methods~\cite{lipson2021raft,li2022practical,xu2023iterative,zhao2023high,wang2021pvstereo} use 
convolutional GRU~\cite{cho2014learning} or LSTM~\cite{graves2012long} as 
the core unit of the update operator to retrieve features from the cost volume and update the 
disparity map, thus avoiding computationally expensive cost aggregation operations. 
Such methods have achieved an overall lead in performance and 
efficiency over other methods and have become the mainstream of research in recent years.

\begin{figure}[t]
  \centering
  \includegraphics[width=\textwidth]{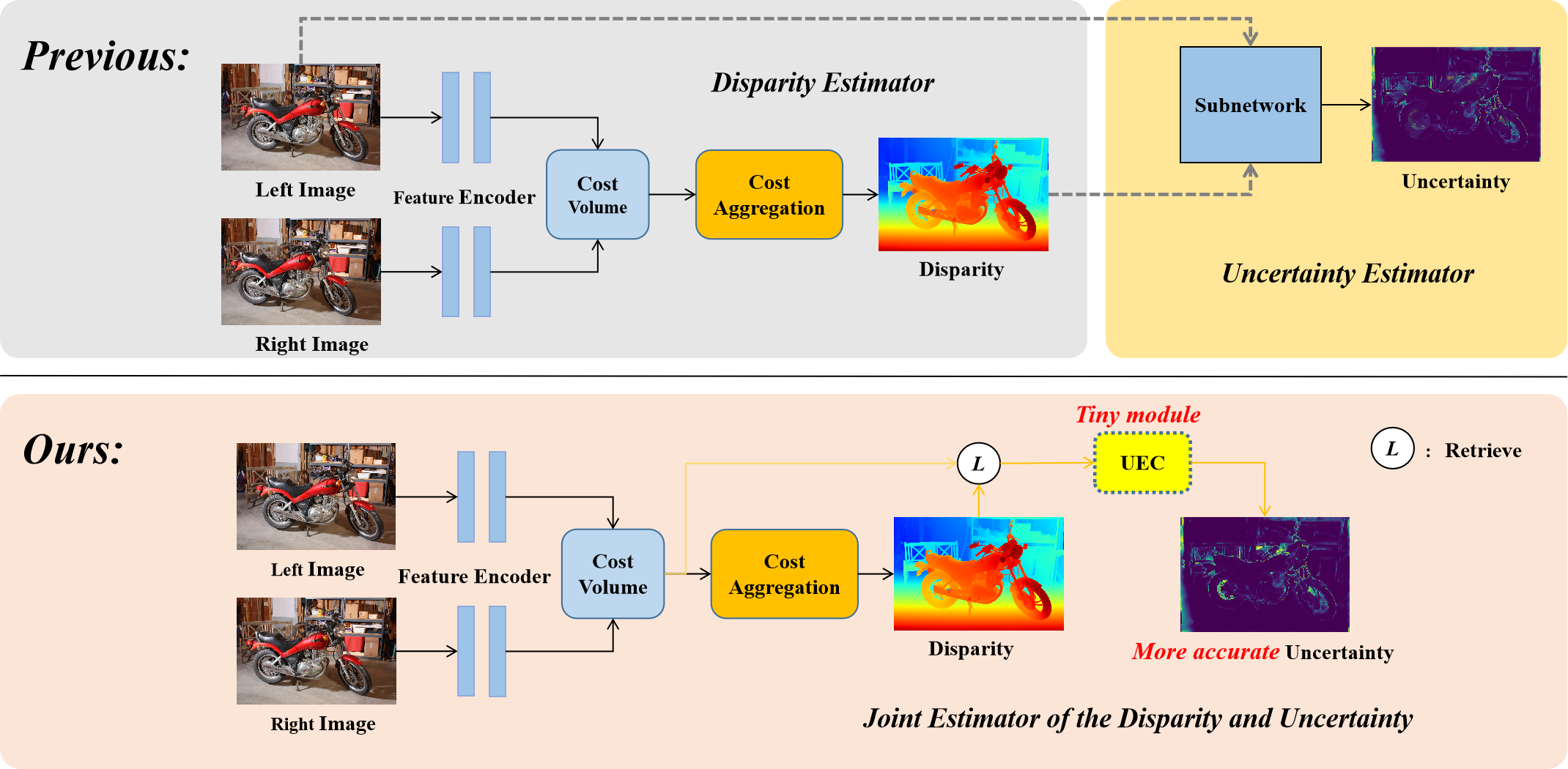}
  \caption{The cost volume-based disparity uncertainty estimation. 
  This figure compares the architecture between the previous work and ours. 
  The previous work only utilises information from the left image. 
  Our work makes full use of the information in the image pairs and 
  avoids redundant feature extraction steps.}
  \label{fig:cdue}
\end{figure}

On the other hand, some works~\cite{shen2021cfnet,chen2023learning,shen2023digging,kim2018unified,shaked2017improved} focus on disparity uncertainty 
estimation, which aims to assist disparity estimation through uncertainty.
UCFNet~\cite{shen2023digging} screens the estimated disparity of a new domain based on uncertainty, 
and uses the screened sparse disparity maps as pseudo-labels to adapt the 
pre-trained model to the new domain. SEDNet~\cite{chen2023learning} proposes a sub-network to 
perform the disparity uncertainty estimation, and uses multi-task 
learning to improve the performance of the disparity estimation network.

Existing studies~\cite{shen2021cfnet,chen2023learning,shen2023digging}, however, essentially use two completely separate networks 
to perform disparity uncertainty estimation and disparity estimation, respectively. 
The uncertainty estimation network takes a single image and the 
corresponding disparity map as as input and directly regresses 
the uncertainty map. This task-separated approach raises the 
computational cost and complexity of the overall architecture. 
Therefore, we investigate the architecture for joint disparity 
and uncertainty estimation. Inspired by the iteration-based 
methods~\cite{lipson2021raft,li2022practical,xu2023iterative,zhao2023high}, 
we recognise that the features of the cost-volume indexed 
by the disparity map contain information about the similarity of the 
left and right maps under the current disparity in terms of context 
and local details, which is an important basis for the uncertainty 
estimation of the disparity map.

In this paper, we propose a new uncertainty estimation method 
named  \emph{Cost volume-based disparity Uncertainty Estimation} (UEC). 
Based on the rich context and local matching information in the 
cost volume, UEC accurately performs the disparity uncertainty 
estimation at a very low computational cost without introducing 
redundant sub-networks (Fig.~\ref{fig:cdue}).  In addition, since the construction of the cost 
volume is a key step in stereo matching, UEC can be inserted into 
almost all stereo matching methods to efficiently complement the 
disparity uncertainty information of predicted disparity.

Based on UEC, we propose two new methods for uncertainty-assisted 
disparity estimation,  \emph{Uncertainty-based Disparity Rectification} (UDR)
and  \emph{Uncertainty-based Disparity update Conditioning} (UDC). The UDR is 
a lightweight disparity updating unit that updates the disparity by the 
change in uncertainty after fine-tuning.

The essence of iterative disparity optimisation is the regression 
of the amount of disparity update on the difference between the current disparity 
and the ground truth. However, the vast majority of 
pixel-by-pixel disparity errors during training and inference 
are within 3 pixels (Table~\ref{tab:LT}), i.e., the ideal distribution of disparity 
updates is a long-tailed distribution. The usual idea 
to mitigate a long-tailed distribution is to increase the weight 
of the tail target in the overall loss function, but this does 
not apply to the amount of disparity updates, for which head 
accuracy is more important. To solve this problem, UDC splits 
the large disparity update into several small disparity updates 
based on the disparity uncertainty. This splitting effectively 
reduces the regression difficulty of the amount of disparity updates, 
thus improving the overall accuracy. In addition, the range of the 
split disparity updates is stable over different domains, which 
contributes to the generalisation performance of the model.

\begin{table*}[h]
  \small
  \caption{Quantitative results for the distribution of 
  disparity updates and the distribution of errors. We pre-train IGEV-Stereo 
  on Scene Flow and conduct experiments directly on the Middlebury 
  2014 training set to statistically characterize the distribution of 
  the amount of disparity updates and disparity errors during the iterations.}
  \label{tab:LT}
  \centering
  \begin{tabular}{lc|cccccccc}
      \toprule
      Value & &	& $<=1px$ & & $(1px, 3px]$	& & $(3px, 5px]$ & & $>5px$ \\
      \midrule
      Updates & & & 97.87\% & & 1.50\%	& & 0.28\% & & 0.35\% \\
      \midrule
      Errors & & & 83.75\% & & 8.20\%	& & 2.48\% & & 5.57\% \\
      \bottomrule
  \end{tabular}
\end{table*}

To further improve the accuracy of small disparity updates, 
we propose \emph{Disparity Rectification loss} (DR loss). 
We construct a dynamic weight that increases the 
focus on pixels with small errors. As the disparity error 
decreases through iterative updates, the increased accuracy of 
small disparity updates contributes to the final disparity 
becoming more accurate. This indicates that DR loss is a 
more advanced and generalised loss function that generally 
improves the performance of iteration-based methods.
We insert UDR, UDC into the iteration-based method and use DR loss during training. 
We name this architecture DR-Stereo, for \emph{Disparity Rectification Stereo}.

In summary, our main contributions are:
\begin{itemize}
\item A novel uncertainty estimation method, UEC, which efficiently 
achieves joint estimation of the disparity and uncertainty based on the cost volume. 
\item Two novel uncertainty-assisted methods for disparity estimation, UDR and UDC. 
The former performs targeted optimisation of the disparity map 
through the uncertainty, while the latter effectively mitigates 
the long-tailed distribution of the amount of disparity updates 
for the iteration-based methods. 
\item An advanced and general loss function, DR loss, 
which increases the focus on small error pixels to improve
 the accuracy of the final disparity. 
\item We propose a new stereo method, DR-Stereo, 
which achieves competitive performance on SceneFlow , 
KITTI benchmarks, Middlebury 2014 and ETH3D.
\end{itemize}

\section{Related Work}

\subsection{Iterative-based Methods}

Inspired by the optical flow network RAFT~\cite{teed2020raft}, 
many iterative-based stereo 
networks~\cite{lipson2021raft,li2022practical,xu2023iterative,zhao2023high,ma2022multiview} have 
been successful in terms of quality and efficiency of disparity 
estimation. RAFT-Stereo~\cite{lipson2021raft} is the first iterative-based 
stereo architecture to be proposed. 
The overall design is based on RAFT~\cite{teed2020raft}, 
replacing the all-pairs of 4D correlation volume with a 3D volume. In addition, 
it introduces a multilevel GRU unit~\cite{cho2014learning}, which remains 
hidden at multiple resolutions with cross connectivity, but still 
generates a single high-resolution disparity update. 
CREStereo~\cite{li2022practical} designs a hierarchical network with 
recurrent refinement, updating the disparity in a coarse-to-fine pattern, 
which leads to a better restoration of fine depth details. 
DLNR~\cite{zhao2023high} proposes an LSTM-based decoupling module to 
iteratively update the disparity and allows features containing 
fine details to be shifted iteratively, mitigating the problem 
that information can be lost during iteration. IGEV-Stereo~\cite{xu2023iterative} 
constructs a combined geometric encoding volume that encodes 
geometric and context information as well as local matching 
details and iteratively indexes them to update the disparity map 
for optimal performance.

\subsection{Estimation of Disparity Uncertainty}

High-performance stereo methods are not error-free, thus it is vital 
to correlate uncertainty with its estimation. UCFNet~\cite{shen2023digging} 
uses pixel-level and region-level uncertainty estimation to filter 
out highly uncertain pixels from the predicted disparity maps 
and generates sparse and reliable pseudo-labels, which is used to 
fine-tune the model so that the model applies to new domains. 
SEDNet~\cite{chen2023learning} proposes a new loss 
function and an uncertainty estimation subnetwork for joint disparity 
and uncertainty estimation, which improves the performance of all 
tasks through multi-task learning. However, all these methods~\cite{shen2021cfnet,chen2023learning,shen2023digging,kim2018unified,kim2020adversarial} 
require both the disparity and the original image as inputs to 
estimate the uncertainty, which leads to inefficiency and redundancy 
in the overall process. We directly predict the disparity 
uncertainty by indexing features of the cost volume by the 
disparity. Our proposed UEC can be integrated into existing 
stereo methods to efficiently estimate the disparity uncertainty 
and efficiently achieves joint estimation of the disparity and uncertainty. 
In addition, by virtue of the low computational 
cost of UEC, we propose two novel uncertainty-assisted methods for disparity estimation, 
UDR and UDC, which further improve the performance of disparity estimation.

\section{Methods}

\begin{figure*}[]
  \centering
  \includegraphics[width=\textwidth]{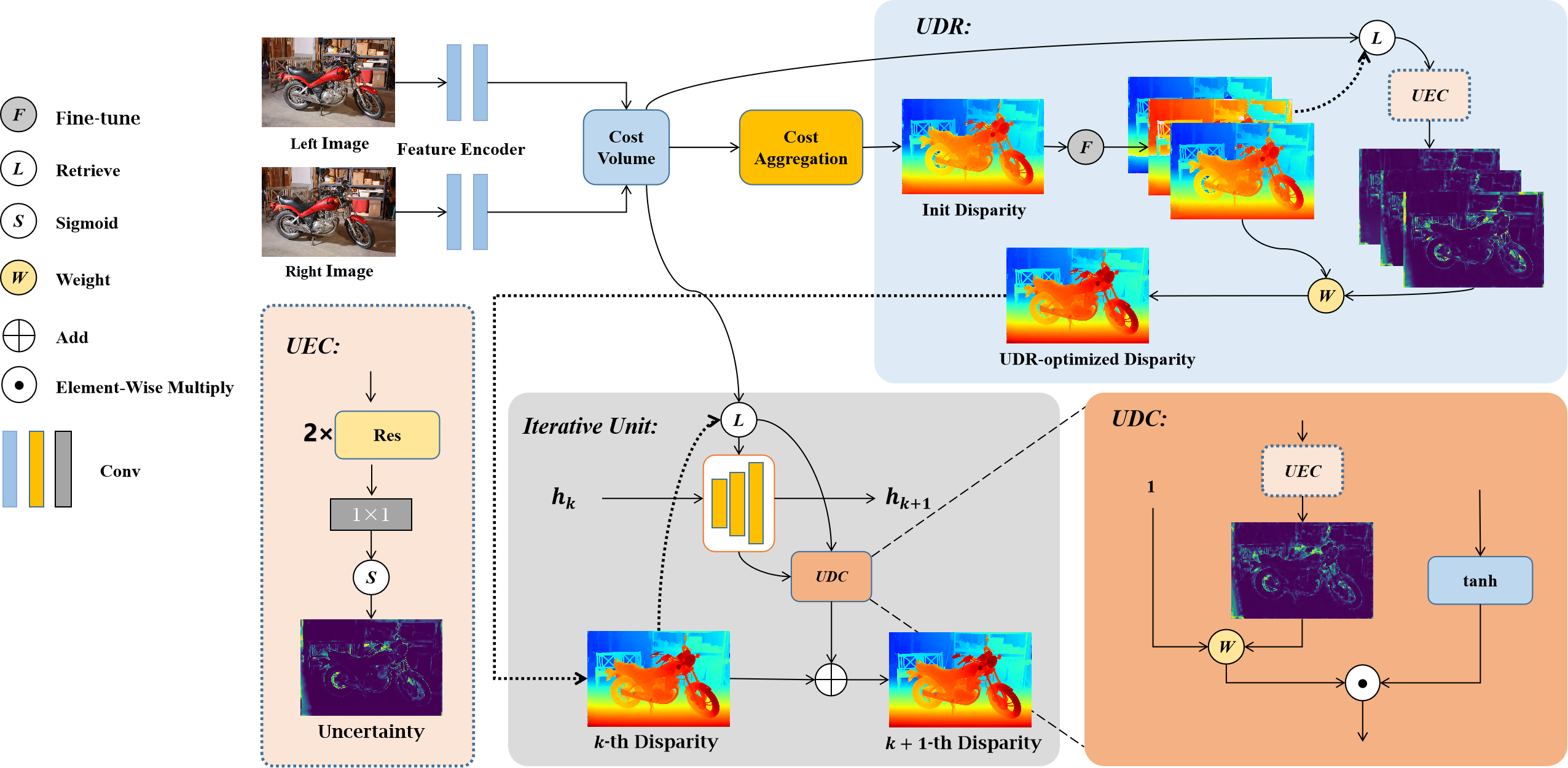}
  \caption{Overview of our proposed DR-Stereo. We estimate the disparity uncertainty 
  by cost volume. The init disparity is coarsely optimised once in the UDR 
  and then finely optimised several times through the iterative unit. 
  In the iterative unit, the proposed UDC moderates the disparity update 
  to keep the update range stable.
  }
  \label{fig:overall}
\end{figure*}

In Fig.~\ref{fig:cdue}, we describe the general process of UEC. In this section, 
we further demonstrate the feasibility of UEC and describe the specific 
implementation of UEC. At the same time, we describe the proposed UDR 
and UDC , and show how to insert them into a general iterative stereo 
matching network (the overall architecture is shown in Fig.~\ref{fig:overall}). 
Finally, we propose DR loss to compute the prediction error for each level of disparity.

\subsection{Cost volume-based Disparity Uncertainty Estimation}

\textbf{Feasibility Study}: To prove the feasibility of the UEC architecture, we compare the loss function of the iteration-based 
method with that of the disparity uncertainty estimation. 
For the former, the loss function can be expressed as:
\begin{align}
  L_{stereo}= \sum_{i=0}^{total_{itr}} \gamma ^{total_{itr}-i}\left \| d_{i} - d_{gt} \right \|
\end{align}
where $d_{gt}$ is the ground truth disparity, $d_{i}$ is the disparity of 
the $i$-th iteration, $total_{itr}$ is the total number of iterations. 
In the iteration-based method, $d_{i}$ is obtained 
by continuous iterative optimisation of the initial disparity map $d_{0}$:
\begin{align}
  d_{i} = d_{0} + \Delta d_{0} + \Delta d_{1} +\dots + \Delta d_{i-1}
\end{align}
where $\Delta d_{i-1}$ is the amount of disparity update in the $i$-th iteration. 
Therefore, the regression target of the amount of disparity update is 
essentially the difference $d_{gt} - d_{pred}$ between the ground truth and the current 
disparity. As for the disparity uncertainty estimation, existing studies usually 
obtain the uncertainty ground truth based on the difference between the 
ground truth of the disparity and the predicted disparity:
\begin{align}
  \label{eq1}
  U_{gt}(d_{pred})=\left\{\begin{matrix}
    0 & , & \left | d_{gt} - d_{pred} \right | \le thr \\
    1 & , & otherwise
  \end{matrix}\right. 
\end{align}

The target of the regression of disparity uncertainty can be 
regarded as a nonlinear transform of the difference $d_{gt} - d_{pred}$ between 
the ground truth of the disparity and the current 
disparity, i.e., a nonlinear transform of the ideal 
amount of disparity update $\Delta d$. Thus estimation of the 
disparity uncertainty by means of the features 
indexed by the disparity map to the cost volume is 
undoubtedly a more reliable and direct way, and this 
architecture is an efficient implementation of a 
joint estimation of disparity and uncertainty.

\textbf{Specific implementation}: The iteration-based methods use the current disparity $d_{k}$
to index features for disparity updating from a two-level $C_{volume}$ pyramid via linear interpolation:
\begin{align}
    G_{f}(d_{k}) =\sum_{i=-r}^{r} Concat\left \{ C_{volume}(d_{k}+i),C_{volume}^{p} (d_{k}/2+i) \right \} 
\end{align}%
where $r$ is the index radius and $p$ denotes the pooling operation. 
Based on the rich similarity information
in $C_{volume}$, the UEC predicts the uncertainty of the current 
disparity using only the retrieved features $G_{f}(d_{k})$ as inputs:
\begin{align}
    U(d_{k}) = \sigma (conv_{1\times 1}(Res(Res(G_{f}(d_{k})))) )
\end{align}%
where $\sigma$ is the sigmoid function and $Res$ is the residual block~\cite{he2016deep}. 
The UEC can estimate the uncertainty of multiple disparity 
maps output from the network at a low computational cost.

\textbf{The uncertainty ground truth }: In order to calculate the disparity 
uncertainty more accurately, we have improved Eq.\ref{eq1}. 
We propose to use the sigmoid function to calculate the ground truth uncertainty:
\begin{align}
  \label{eq2}
  U_{gt} (d_{k}) = \sigma (a\times \left | d_{gt}-d_{k} \right |-thr)  
\end{align}
where $a$ is the transition distance between correct to incorrect predictions. 
As the $a$ tends to infinity, Eq.\ref{eq2} is approximately equivalent to Eq.\ref{eq1}. 
In Section~\ref{Ablation Study}, we explore the effect of the uncertainty ground truth 
setting on the accuracy of the uncertainty estimation and find that 
the best results are obtained with $a = 1.5$ and $thr = 3.0$.

\subsection{Disparity Rectification}\label{Disparity Rectification method}

In this section, we introduce two uncertainty-assisted methods for disparity estimation, 
\emph{Uncertainty-based Disparity Rectification} (UDR) and \emph{Uncertainty-based 
Disparity update Conditioning} (UDC).

\textbf{Uncertainty-based Disparity Rectification}: The disparity uncertainty is
 related to the magnitude of the expected error. We consider it a beneficial 
 update if the disparity uncertainty is reduced. 
 In \emph{Uncertainty-based Disparity Rectification} (UDR), the disparity map is 
 fine-tuned as a whole and then the uncertainty is recalculated by UEC. 
 The original disparity is updated by combining the recalculated uncertainty 
 with the fine-tuning magnitude:
 \begin{align}
    d_{UDR}^{k+1}  = d_{k}+ s\times (U(d_{k}-s)-U(d_{k}+s))
\end{align}%
where $s$ is the fine-tuning magnitude of the disparity map. The UDR 
is an intuitive and effective method of disparity optimization, and 
in this paper we use it to optimize the initial disparity map, which 
has the greatest impact on the disparity performance.

\textbf{Uncertainty-based Disparity Update Conditioning}: The process of 
updating the disparity for the iterative stereo matching method can be 
described as follows:
\begin{align}
    h_{k}  = Unit_{update}(G_{f}(d_{k}), h_{k-1},d_{k})
\end{align}%
\begin{align}
    d_{k+1}   = d_{k} + Decoder_{d} (h_{k}) 
\end{align}%
where $Unit_{update}$ is the disparity update unit, $h_{k}$ is the current 
hidden state of the update unit, $Decoder_{d}$ is the decoder used to output the amount of disparity updates..
We propose \emph{Uncertainty-based Disparity update Conditioning} (UDC), which regulates 
the amount of disparity update through the disparity uncertainty:
\begin{align}
    d_{UDC}^{k+1}   = d_{k} + m\times tanh(\frac{Decoder_{d} (h_{k})}{m})\odot (1+0.5\times U(d_{k}))  
\end{align}
where $m$ is the modulation factor of the UDC and $\odot$ denotes the Hadamard Product. 
We control the upper limit of the amount of disparity updates by $m$, 
thus splitting the large amount of disparity updates into several smaller ones. 
In addition, the $h_{k}$ contains information $\left \{ G_{f}(d_{i}) \right \} _{i=1}^{k} $ about 
the previous disparity, providing additional guidance on the disparity update. 
For pixels with high uncertainty, we encourage 
larger disparity updates so that the $h_{k}$ receives more surrounding features.

Notably, the UDC and UDR can be incorporated repeatedly into iteration-based 
methods at almost negligible cost (see Section~\ref{Ablation Study} for more details).

\subsection{Disparity Rectification Loss}

We observe that in the existing loss functions (e.g., L1 loss and Smooth L1 loss), 
the effect of pixels decreases as the error decreases, 
which restricts the upper limit of the accuracy of small disparity updates. 
Therefore, we propose weights that tend to focus on pixels with small errors:
\begin{align}
    w_{DR}(d_{k}) = exp(-\alpha \times \left | d_{k}-d_{gt} \right |) +\beta  
\end{align}%
where $\alpha$ and $\beta$ are used to control the distribution and the overall scale of weights. 
It is shown in Section~\ref{Loss Function} how to combine 
this weight with L1 loss and Smooth L1 loss. In this paper, 
we set $\alpha$ to 1/8 and $\beta$ to 1/10, and demonstrate its validity 
experimentally (see Section~\ref{Ablation Study} and~\ref{Benchmarks} for more details).

\subsection{Loss Function}\label{Loss Function}

We calculate the smooth L1 loss on the initial disparity $d_{0}$ and the
UDR-optimized disparity $d_{UDR}^{1}$ and weight them 
using their rectification weights:
\begin{align}
    L_{init} = w_{DR}(d_{0})\odot Smooth_{L_{1} } (d_{0}-d_{gt})  
\end{align}%
\begin{align}
    L_{UDR} = w_{DR}(d_{UDR}^{1})\odot Smooth_{L_{1} } (d_{UDR}^{1}-d_{gt})  
\end{align}

We calculate the L1 loss on all predicted disparities $\left \{ d_{UDC}^{i}  \right \}_{i=1}^{total_{itr}}$ and jointly 
weight them using exponentially increasing weights and rectification weights:
\begin{align}
    L_{UDC}=\sum_{i=1}^{total_{itr}} \gamma ^{total_{itr}-i} \times w_{DR}(d_{UDC}^{i})\odot \left \| d_{UDC}^{i} - d_{gt} \right \|  
\end{align}
where $\gamma$ = 0.9. We calculate the smooth L1 loss on all disparity uncertainties:
\begin{align}
    L_{UEC}(d_{UDC}^{i}) = Smooth_{L_{1} } (U(d_{UDC}^{i})-U_{gt} (d_{UDC}^{i}))  
\end{align}

The total loss is defined as:
\begin{align}
    L_{total} =  L_{init} + L_{UDR} + L_{UDC} + \sum_{i=1}^{total_{itr}} L_{UEC}(d_{UDC}^{i})
\end{align}

\section{Experiments}

\subsection{DATASETS}

\textbf{Scene Flow}~\cite{mayer2016large} is a synthetic dataset containing 
35,454 training pairs and 4,370 test pairs, and we use Finalpass 
of Scene Flow because it is closer to real-world images. 
\textbf{KITTI 2012}~\cite{geiger2012we} and \textbf{KITTI 2015}~\cite{menze2015object} are 
datasets of real-world driving scenes. KITTI 2012 contains 194 training pairs 
and 195 test pairs, while KITTI 2015 contains 200 training 
pairs and 200 test pairs. Both KITTI datasets provide sparse 
ground truth disparities obtained using LiDAR. \textbf{Middlebury 2014}~\cite{scharstein2014high} 
is an indoor dataset that provides 15 training pairs and 15 test 
pairs, with some of the samples being inconsistent under lighting 
or color conditions. \textbf{ETH3D}~\cite{schops2017multi} contains a variety of indoor and outdoor 
scenes, and provides 27 training pairs and 20 test pairs.

\subsection{Implementation Details}

The framework is implemented using PyTorch and 
We used NVIDIA RTX 3090 GPUs for our experiments. On Scene Flow, the final model 
is trained with a batch size of 8 for a total of 200k steps, 
while the ablation experiments are trained with a batch size 
of 4 for 50k steps. On KITTI, we finetune the pre-trained 
sceneflow model on a mixed KITTI 2012 and KITTI 2015 training 
image pair for 50k steps. The ablation experiments are trained 
using 10 update iterations during training, and the final model 
is trained using 22 update iterations. The final model and 
the ablation experiments use a one-cycle learning rate 
schedule with a learning rate of 0.0002 and 0.0001, 
respectively. We demonstrate the generalization performance 
of our method by testing the pre-trained Scene 
Flow model directly on the training sets of Middlebury 2014 and ETH3D.

\subsection{Estimation of Disparity Uncertainty}\label{UEC}

\begin{figure}[t]
  \centering
  \includegraphics[width=\textwidth]{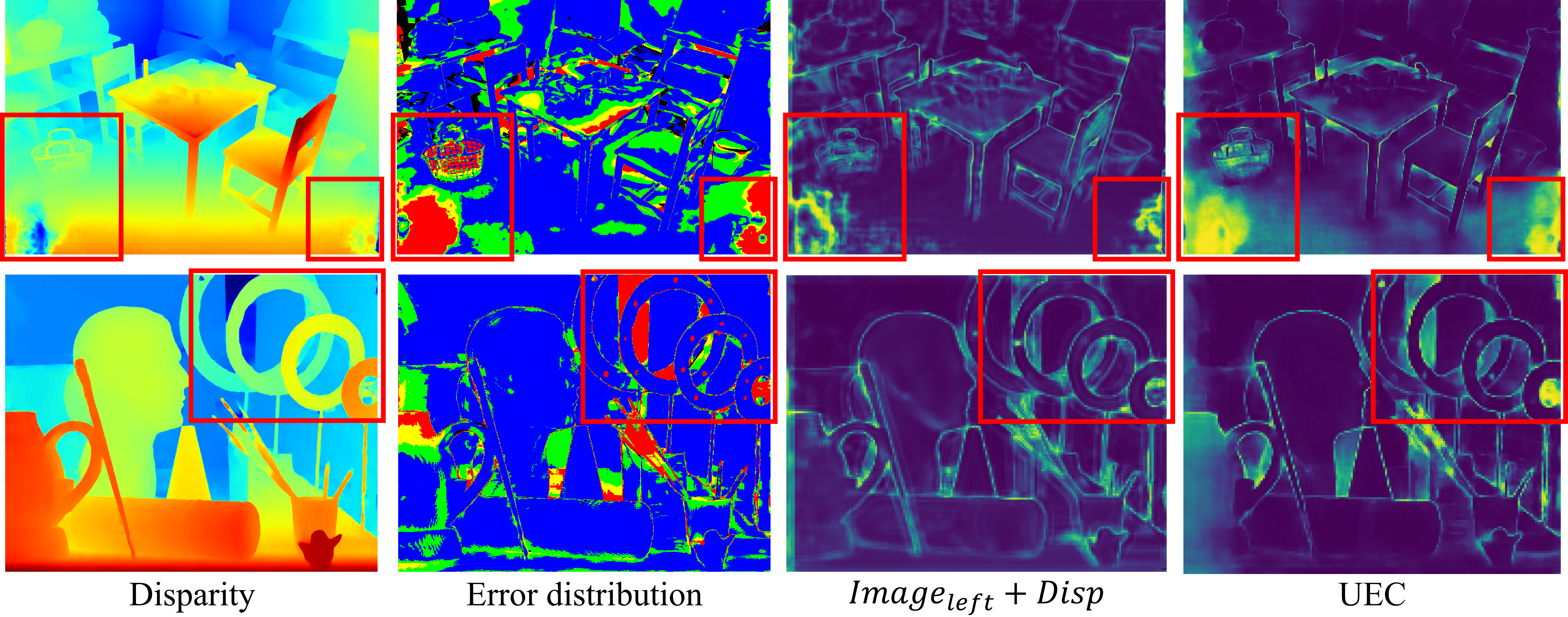}
  \caption{The qualitative results of UEC on Middlebury 2014.
  The error distribution of disparity is plotted with the largest error in 
  the red region and the smallest error in the blue region.
  We pre-train our model on Scene Flow and test it directly on Middlebury 2014.
  On the new domain, the sensitivity of UEC to disparity error 
  is superior to that of task-separated architectures.
  }
  \label{fig:duc}
\end{figure}

In this section, we compare the uncertainty estimation performance 
between the UEC architecture and the task-separated architecture for the 
same number of parameters and explore the impact of the ground truth setting.
To confirm the generality of the UEC architecture, we conducted experiments 
with several common cost volumes. Comparing (a) and others in Table~\ref{tab:tab_uec}, 
the uncertainty estimation performance of our proposed UEC architectures 
generally outperforms that of task-separated architectures. 
Among all experiments of UEC architectures, the combined volume Gwc8-Cat8 
achieves the best overall performance, while Correlation performs 
relatively poorly due to the loss of too much information during the 
construction process (but still approximates the task-separated architecture).
In Table~\ref{tab:tab_uec}, the improvement from (c) to (d) and from (i) to (j) 
is significant, which indicates that richer similarity information 
is beneficial for uncertainty estimation. This further exemplifies 
the commonality between uncertainty uncertainty estimation and disparity 
estimation in terms of required features. In addition, with smoother 
ground truth settings ((f)$\sim $(i)), the UEC achieves an overall 
improvement on the uncertainty estimation performance.
Fig.~\ref{fig:duc} shows the qualitative results of the two uncertainty 
estimation architectures at Middlebury 2014. The proposed UEC significantly 
outperforms the previous architecture in terms of generalisation 
for disparity uncertainty estimation and performs well in the object edge region.

\begin{table}[t]
  \small
  \caption{Quantitative results of uncertainty estimation on the SceneFlow test set. 
  In addition to using the area under the ROC curve (AUC), 
  we propose per-pixel uncertainty error (PUE) to quantitatively 
  evaluate uncertainty estimation performance. 
  Gwc refers to Group-wise correlation volume, and Cat stands for Concatenation volume. \textbf{Bold}: Best.
  }
  \label{tab:tab_uec}
  \centering
  \begin{tabular}{c|l|cc|cc||c|l|cc|cc}
      \toprule
      & Input &	$a$ & $thr$	& AUC$\downarrow $ & PUE$\downarrow $ & & Input &	$a$ & $thr$	& AUC$\downarrow $ & PUE$\downarrow $ \\
      \midrule
      (a) & $Image_{left}+Disp$ & 1.5 & 3.0 & 0.155 & 0.061 & (f) & Gwc8 & 1000.0 & 2.0	& 0.150 & \cellcolor{gray!20}0.044 \\
      (b) & Correlation & 1.5 & 3.0	& 0.154 & 0.059 & (g) & Gwc8 & 1.0 & 4.0	& 0.146 & \textbf{0.041} \\
      (c) & Cat8 & 1.5 & 3.0	& 0.147 & 0.048 & (h) & Gwc8 & 2.0 & 2.0	& 0.142 & 0.062 \\
      (d) & Cat16 & 1.5 & 3.0	& \textbf{0.133} & 0.045 & (i) & Gwc8 & 1.5 & 3.0	& 0.144 & 0.047 \\
      (e) & Gwc8-Cat8 & 1.5 & 3.0	& \textbf{0.133} & \cellcolor{gray!20}0.044 & (j) & Gwc16 & 1.5 & 3.0	& \cellcolor{gray!20}0.135 & 0.046 \\
      \bottomrule
  \end{tabular}
\end{table}

\subsection{Ablation Study}\label{Ablation Study}

\begin{table}
  \small 
  \caption{Ablation study and complexity of DR-stereo. 
  The baseline is IGEV-Stereo. The last two columns are 
  the results when the size of input image is $1248\times384$. \textbf{Bold}: Best.}
  \label{tab:CPM}
  \centering
    \begin{tabular}{l|ccc|cc}
      \toprule
      Methods & Scene Flow & Middlebury-H & ETH3D & Params(M) & Time(s) \\
      \midrule
      Baseline & 3.65 & 8.44 &  4.49 & 12.60 & 0.155 \\
      \midrule
      UDC (C.) + DR loss       & 3.37 & 6.98 & 4.35 & 12.77 & 0.158 \\
      UDR (R.) + DR loss       & \cellcolor{gray!20}3.34 & \cellcolor{gray!20}6.64 & \cellcolor{gray!20}4.05 & 12.77 & 0.160 \\
      C.+R.+DR loss (DR-Stereo) & \textbf{3.33} & \textbf{6.39} & \textbf{3.99} & 12.77 & 0.161 \\
      \bottomrule
  \end{tabular}
\end{table}

\textbf{Combinations of proposed methods}: 
As shown in Table~\ref{tab:CPM}, we experiment with multiple 
combinations of the proposed methods. 
Comparing the results with those in Tables~\ref{tab:UDCUDR} and \ref{tab:DRLOSS}, 
our methods achieve further improvements after combination. 
In addition, we record the increase in the number of parameters, 
time consumption after inserting the design modules. 
It can be seen that inserting the UDR and UDC puts little 
burden for the model. As we present in Section~\ref{Disparity Rectification method}, 
the combination of UDR and UDC does not increase the 
number of parameters compared to a single method, 
since they share parameters from the UEC module.

\subsection{Update on Bad Initial Disparity}\label{baddisp}
In this section, we investigate the role of UDC in regulating 
the process of disparity update. The disparity distributions 
of different datasets tend to be highly different, which leads 
to large initial disparity errors (i.e., the amount of ideal 
disparity update) on new domains for iteration-based methods. 
Our proposed UDC splits the large disparity update into several 
small disparity updates, which alleviates the difference in the 
distribution of ideal disparity updates between different domains. 
Fig.~\ref{fig:UDC} shows the effect of UDC on the disparity update process 
in extreme cases. In regions with large initial disparity errors, 
the method using UDC performs a faster stepwise optimisation 
of the disparity and generates a more accurate disparity 
with the same number of iterations.

\begin{figure}[t]
  \centering
  \includegraphics[width=\textwidth]{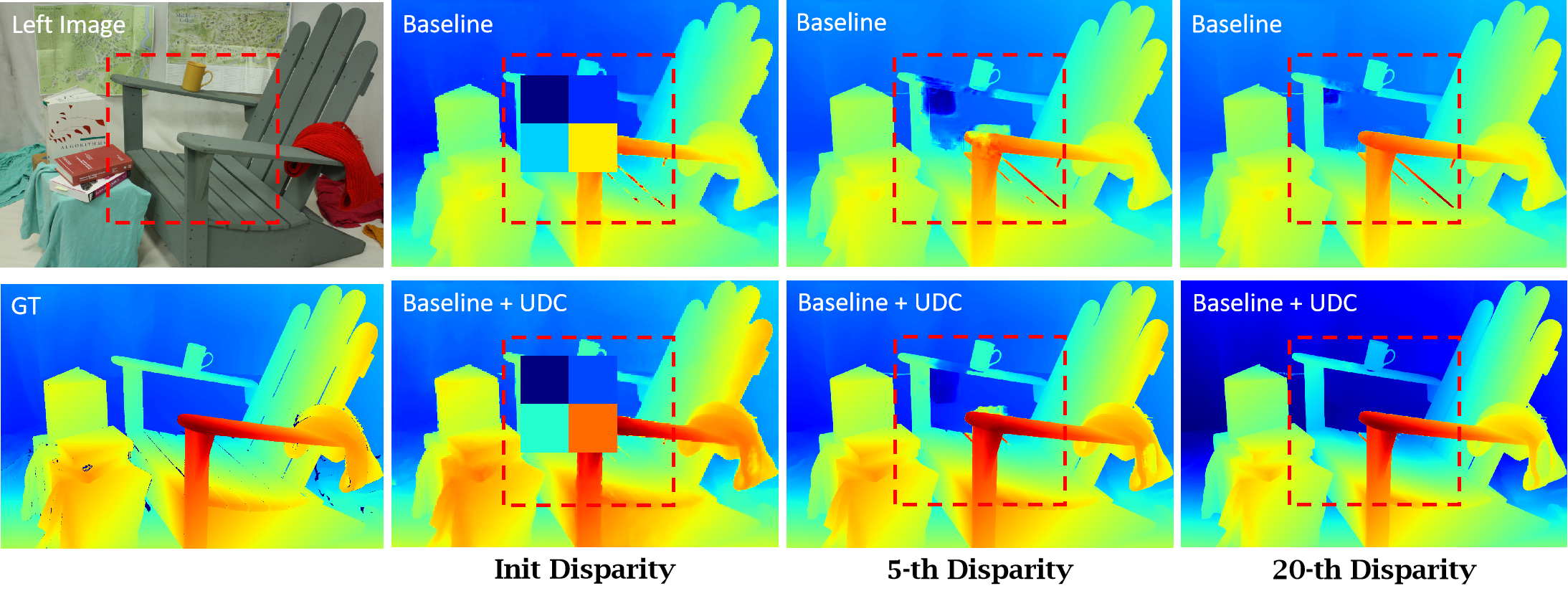}
  \caption{
    The effect of UDC in the disparity update process. 
    We mosaic over the initial disparity to simulate the 
    disparity update process in extreme cases. 
    Test image from Middlebury 2014. The baseline is IGEV-Stereo.
  }
  \label{fig:UDC}
\end{figure}

\subsection{Benchmarks}\label{Benchmarks}

In this section, we compare DR-Stereo with 
the state-of-the-art methods published on Scene Flow and KITTI.
Tables~\ref{tab:Benchmarks} shows the quantitative results. 
With similar training strategies, DR-Stereo achieves a new SOTA EPE on the Scene Flow test set. 
Evaluation results on the KITTI benchmark show that DR-Stereo achieves 
the best performance on the vast majority of metrics. At the time of writing, 
our method outperforms all published methods on the online KITTI 2015 leaderboard.

\begin{table}[t]
  \caption{Quantitative evaluation on Scene Flow and KITTI 2015. \textbf{Bold}: Best.}
  \label{tab:Benchmarks}
  \centering
  \begin{tabular}{l|c|ccc}
      \toprule
      \multirow{2}{*}{Method}  & \multirow{2}{*}{Scene Flow} & \multicolumn{3}{c}{KITTI 2015} \\
        & & D1-bg & D1-fg & D1-all \\
      \midrule
      CREStereo~\cite{li2022practical} & - & 1.45 & 2.86 & 1.69   \\
      DLNR~\cite{zhao2023high} & 0.48 & \textbf{1.37} & 2.59 & 1.76 \\
      Croco-Stereo~\cite{weinzaepfel2023croco} & - & \cellcolor{gray!20}1.38 & \cellcolor{gray!20}2.65 & \cellcolor{gray!20}1.59 \\
      UPFNet~\cite{shen2023digging} & - & 1.38 & 2.85 & 1.62  \\
      PSMNet~\cite{chang2018pyramid} & 1.09 & 1.86 & 4.62 & 2.32  \\
      GANNet~\cite{zhang2019ga} & 0.80 & 1.48 & 3.46 & 1.81   \\
      GwcNet~\cite{guo2019group} & 0.98 & 1.74 & 3.93 & 2.11  \\
      AcfNet~\cite{zhang2020adaptive} & 0.87 & 1.51 & 3.80 & 1.89  \\
      ACVNet~\cite{xu2022attention} & 0.48 & \textbf{1.37} & 3.07 & 1.65 \\
      RAFT-Stereo~\cite{lipson2021raft} & 0.56 & 1.58 & 3.05 & 1.82   \\
      IGEV-Stereo~\cite{xu2023iterative} & \cellcolor{gray!20}0.47 & 1.38 & 2.67 & \cellcolor{gray!20}1.59 \\
      \midrule
      DR-Stereo(ours) & \textbf{0.45} & \textbf{1.37} & \textbf{2.50} & \textbf{1.56}\\
      \bottomrule
  \end{tabular}
\end{table}

\subsection{Zero-shot Generalization}\label{Zeroshot}
We pre-train DR-Stereo on Scene Flow and test it directly on Middlebury 2014 and ETH3D.
As shown in Table~\ref{tab:zero}, Our DR-Stereo achieves very competitive generalisation performance. 
Compared with the original optimal method IGEV-Stereo, our method achieves an 
overall improvement. 

\begin{table}[h]
  \small
  \centering
  \caption{Synthetic data generalization experiments. 
  We pre-train our model on Scene Flow and test it directly on Middlebury 2014 and ETH3D.
  The 2-pixel error rate is used for Middlebury 2014, and 1-pixel error rate for ETH3D. \textbf{Bold}: Best.}
  \label{tab:zero}
  \begin{tabular}{l|cc|c||l|cc|c}
      \toprule
      \multirow{2}{*}{Method} & \multicolumn{2}{c}{Middlebury} & \multirow{2}{*}{ETH3D} & \multirow{2}{*}{Method} & \multicolumn{2}{c}{Middlebury} & \multirow{2}{*}{ETH3D}  \\
       & half & quarter & &  & half & quarter &\\
      \midrule
      PSMNet~\cite{chang2018pyramid}  & 15.8 & 9.8 & 10.2 & CFNet~\cite{shen2021cfnet} & 15.3 & 9.8 & 5.8 \\
      GANNet~\cite{zhang2019ga} & 13.5 & 8.5 & 6.5 & RAFT-Stereo~\cite{lipson2021raft} &  8.7 & 7.3 & \textbf{3.2} \\ 
      DSMNet~\cite{zhang2020domain} & 13.8 & 8.1 & 6.2 & IGEV-Stereo~\cite{xu2023iterative} &  \cellcolor{gray!20}7.1 & \cellcolor{gray!20}6.2 & 3.6 \\ 
      STTR~\cite{li2021revisiting} & 15.5 & 9.7 & 17.2 & DR-Stereo(ours) & \textbf{5.5} & \textbf{5.2} & \cellcolor{gray!20}3.5 \\
      \bottomrule
  \end{tabular}
\end{table}

\section{Conclusion}

In this paper, we propose \emph{Cost volume-based disparity Uncertainty Estimation} (UEC). Based on the rich feature information 
in the cost volume, UEC accurately estimates the disparity 
uncertainty with very low computational cost. On this basis, 
we propose \emph{Uncertainty-based Disparity Rectification} (UDR) and 
\emph{Uncertainty-based Disparity update Conditioning} (UDC). These two 
methods significantly improve the generalisation performance 
of the iteration-based methods. We propose the \emph{Disparity 
Rectification loss} (DR loss), which improves the accuracy of the small 
amount of disparity updates. This improvement contributes to the 
final disparity becoming more accurate. Finally, we insert UDR, UDC, 
and DR loss into the iteration-based method and name this new method 
\emph{Disparity Rectification Stereo} (DR-Stereo). DR-Stereo achieves 
competitive performance on several publicly available datasets.

\section{More results of UDR}

In DR-Stereo, we update the initial disparity map using the 
UDR module, which efficiently corrects for obvious disparity errors. 
Table~\ref{tab:UDR} shows the quantitative results.

\begin{table}
  \small 
  \caption{Quantitative results for a single UDR module.We record the results of the UDR updating of the initial disparity on multiple datasets.}
  \label{tab:UDR}
      \centering
      \begin{tabular}{l|cc|cc|cc}
          \toprule
          \multirow{2}{*}{Experiment} & \multicolumn{2}{c}{Scene Flow} & \multicolumn{2}{c}{Middlebury-H} & \multicolumn{2}{c}{ETH3D} \\
           & EPE(px) & $>3px(\%)$ & EPE(px) & $>2px(\%)$ & EPE(px) & $>1px(\%)$ \\
          \midrule
          Init disparity & 1.04 & 4.96 & 1.60 & 11.87 & 0.76 & 10.38  \\
          + UDR & 0.98 & 4.77 & 1.51 & 11.01 & 0.72 & 9.56  \\
          \bottomrule
      \end{tabular}
  \end{table}

\section{Principle of UDC}

The UDC splits large disparity updates, thus mitigating the imbalance in the distribution of the ideal amount of updates between different datasets.
Fig~\ref{fig:UDC} visualises the impact of the splitting process.

\begin{figure}[h]
  \centering
  \includegraphics[width=0.8\textwidth]{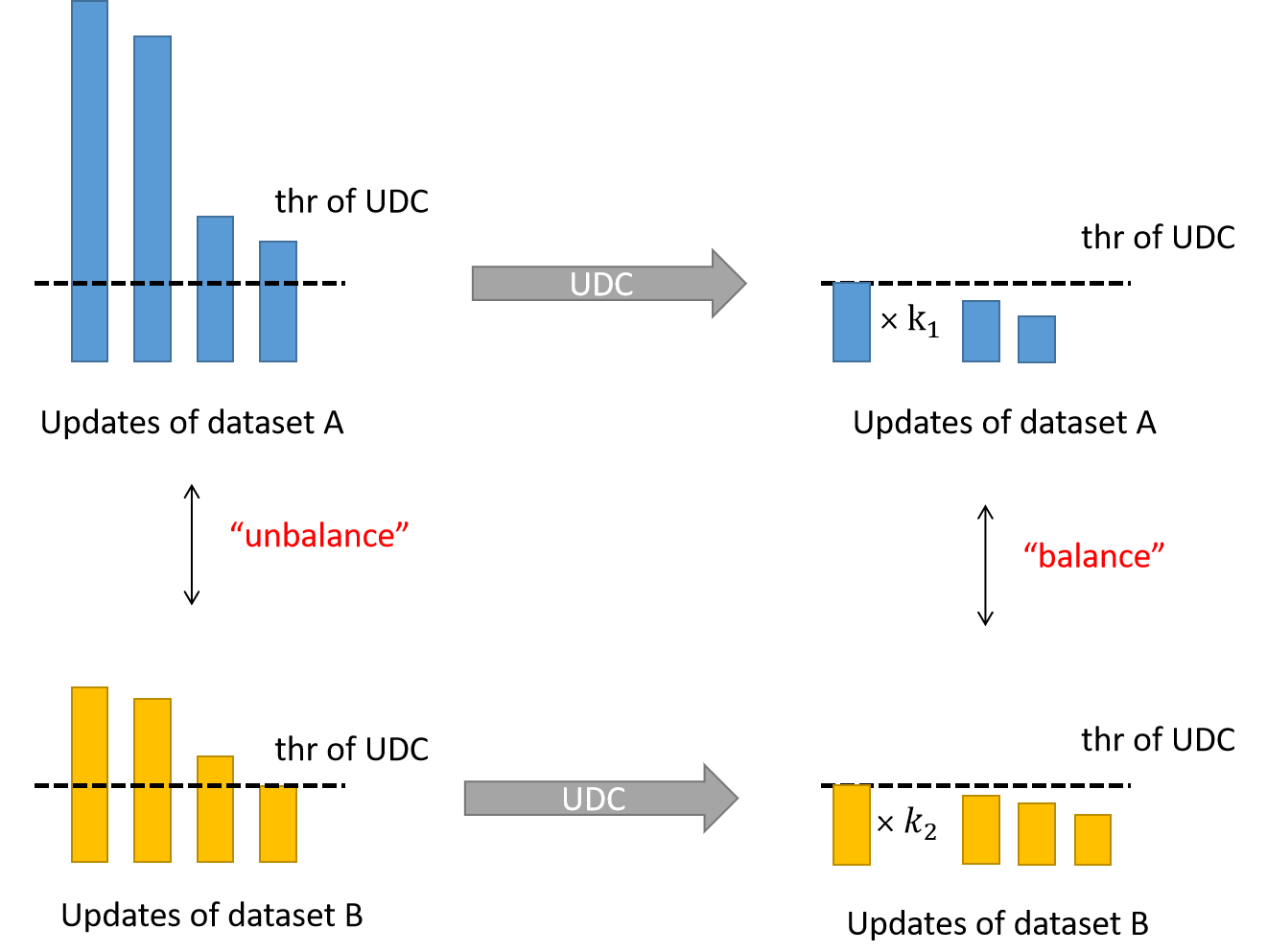}
  \caption{
    Splitting process of UDC on different datasets.
  }
  \label{fig:UDC}
\end{figure}

\section{Other attempts of DR loss}
In this section, we explore alternative forms of DR loss. 
DRloss is a loss function that focuses on small disparity updates. 
Table~\ref{tab:DRLOSS} shows the many forms we have experimented with.
We observe that its specific forms can be varied. 
While the sigmoid form of DR loss achieves more performance gains, 
we find experimentally that it is difficult to train when the weight 
bias is small. Therefore, we did not adopt this setting in the end. 
We will investigate more possibilities of DR loss in the future.

\begin{table}[t]
  \small 
  \caption{Other attempts of DR loss. The baseline is IGEV-Stereo.}
  \label{tab:DRLOSS}
  \centering
    \begin{tabular}{l|cc|cc|c}
        \toprule
        \multirow{2}{*}{Methods}   & \multicolumn{2}{c}{Scene Flow} & \multicolumn{2}{c}{Middlebury-H} & ETH3D \\
           & EPE(px) & $>3px(\%)$ & EPE(px) & $>2px(\%)$ & $>1px(\%)$ \\
      \midrule
      baseline  & 0.72 & 3.65 & 1.28 & 8.44 & 4.49 \\
      \midrule
      $exp(-0.125 \times \left | d_{k}-d_{gt} \right |) +0.1$ & 0.80 & \cellcolor{gray!20}3.26 & \cellcolor{gray!20}1.20 & \cellcolor{gray!20}7.47 & 3.99 \\
      $exp(-0.125 \times \left | d_{k}-d_{gt} \right |) +0.5$ & \textbf{0.71} & 3.37 & 1.27 & 7.79  & 4.11 \\
      \midrule
      $sigmoid(6 - 0.1 \times  \left | d_{k}-d_{gt} \right |)+0.1$ &  \cellcolor{gray!20}0.72& 3.43 & \textbf{1.16} &\textbf{7.30}&\cellcolor{gray!20}3.89\\
      $sigmoid(6 - 0.5 \times  \left | d_{k}-d_{gt} \right |)+0.1$ &  \cellcolor{gray!20}0.72& \textbf{3.25} & 1.25 &8.00&\textbf{3.88}\\
      \bottomrule
  \end{tabular}
\end{table}

%
%
\bibliographystyle{splncs04}
\bibliography{ref}

\begin{thebibliography}{10}
\providecommand{\url}[1]{\texttt{#1}}
\providecommand{\urlprefix}{URL }
\providecommand{\doi}[1]{https://doi.org/#1}

\bibitem{chang2018pyramid}
Chang, J.R., Chen, Y.S.: Pyramid stereo matching network. In: Proceedings of the IEEE conference on computer vision and pattern recognition. pp. 5410--5418 (2018)

\bibitem{chen2023learning}
Chen, L., Wang, W., Mordohai, P.: Learning the distribution of errors in stereo matching for joint disparity and uncertainty estimation. In: Proceedings of the IEEE/CVF Conference on Computer Vision and Pattern Recognition. pp. 17235--17244 (2023)

\bibitem{cho2014learning}
Cho, K., Van~Merri{\"e}nboer, B., Gulcehre, C., Bahdanau, D., Bougares, F., Schwenk, H., Bengio, Y.: Learning phrase representations using rnn encoder-decoder for statistical machine translation. arXiv preprint arXiv:1406.1078  (2014)

\bibitem{duggal2019deeppruner}
Duggal, S., Wang, S., Ma, W.C., Hu, R., Urtasun, R.: Deeppruner: Learning efficient stereo matching via differentiable patchmatch. In: Proceedings of the IEEE/CVF international conference on computer vision. pp. 4384--4393 (2019)

\bibitem{geiger2012we}
Geiger, A., Lenz, P., Urtasun, R.: Are we ready for autonomous driving? the kitti vision benchmark suite. In: 2012 IEEE conference on computer vision and pattern recognition. pp. 3354--3361. IEEE (2012)

\bibitem{graves2012long}
Graves, A., Graves, A.: Long short-term memory. Supervised sequence labelling with recurrent neural networks pp. 37--45 (2012)

\bibitem{gu2020cascade}
Gu, X., Fan, Z., Zhu, S., Dai, Z., Tan, F., Tan, P.: Cascade cost volume for high-resolution multi-view stereo and stereo matching. In: Proceedings of the IEEE/CVF conference on computer vision and pattern recognition. pp. 2495--2504 (2020)

\bibitem{guo2019group}
Guo, X., Yang, K., Yang, W., Wang, X., Li, H.: Group-wise correlation stereo network. In: Proceedings of the IEEE/CVF conference on computer vision and pattern recognition. pp. 3273--3282 (2019)

\bibitem{he2016deep}
He, K., Zhang, X., Ren, S., Sun, J.: Deep residual learning for image recognition. In: Proceedings of the IEEE conference on computer vision and pattern recognition. pp. 770--778 (2016)

\bibitem{kim2018unified}
Kim, S., Min, D., Kim, S., Sohn, K.: Unified confidence estimation networks for robust stereo matching. IEEE Transactions on Image Processing  \textbf{28}(3),  1299--1313 (2018)

\bibitem{kim2020adversarial}
Kim, S., Min, D., Kim, S., Sohn, K.: Adversarial confidence estimation networks for robust stereo matching. IEEE Transactions on Intelligent Transportation Systems  \textbf{22}(11),  6875--6889 (2020)

\bibitem{li2022practical}
Li, J., Wang, P., Xiong, P., Cai, T., Yan, Z., Yang, L., Liu, J., Fan, H., Liu, S.: Practical stereo matching via cascaded recurrent network with adaptive correlation. In: Proceedings of the IEEE/CVF conference on computer vision and pattern recognition. pp. 16263--16272 (2022)

\bibitem{li2021revisiting}
Li, Z., Liu, X., Drenkow, N., Ding, A., Creighton, F.X., Taylor, R.H., Unberath, M.: Revisiting stereo depth estimation from a sequence-to-sequence perspective with transformers. In: Proceedings of the IEEE/CVF international conference on computer vision. pp. 6197--6206 (2021)

\bibitem{liang2019stereo}
Liang, Z., Guo, Y., Feng, Y., Chen, W., Qiao, L., Zhou, L., Zhang, J., Liu, H.: Stereo matching using multi-level cost volume and multi-scale feature constancy. IEEE transactions on pattern analysis and machine intelligence  \textbf{43}(1),  300--315 (2019)

\bibitem{lipson2021raft}
Lipson, L., Teed, Z., Deng, J.: Raft-stereo: Multilevel recurrent field transforms for stereo matching. In: 2021 International Conference on 3D Vision (3DV). pp. 218--227. IEEE (2021)

\bibitem{ma2022multiview}
Ma, Z., Teed, Z., Deng, J.: Multiview stereo with cascaded epipolar raft. In: European Conference on Computer Vision. pp. 734--750. Springer (2022)

\bibitem{mayer2016large}
Mayer, N., Ilg, E., Hausser, P., Fischer, P., Cremers, D., Dosovitskiy, A., Brox, T.: A large dataset to train convolutional networks for disparity, optical flow, and scene flow estimation. In: Proceedings of the IEEE conference on computer vision and pattern recognition. pp. 4040--4048 (2016)

\bibitem{menze2015object}
Menze, M., Geiger, A.: Object scene flow for autonomous vehicles. In: Proceedings of the IEEE conference on computer vision and pattern recognition. pp. 3061--3070 (2015)

\bibitem{scharstein2014high}
Scharstein, D., Hirschm{\"u}ller, H., Kitajima, Y., Krathwohl, G., Ne{\v{s}}i{\'c}, N., Wang, X., Westling, P.: High-resolution stereo datasets with subpixel-accurate ground truth. In: Pattern Recognition: 36th German Conference, GCPR 2014, M{\"u}nster, Germany, September 2-5, 2014, Proceedings 36. pp. 31--42. Springer (2014)

\bibitem{schops2017multi}
Schops, T., Schonberger, J.L., Galliani, S., Sattler, T., Schindler, K., Pollefeys, M., Geiger, A.: A multi-view stereo benchmark with high-resolution images and multi-camera videos. In: Proceedings of the IEEE conference on computer vision and pattern recognition. pp. 3260--3269 (2017)

\bibitem{shaked2017improved}
Shaked, A., Wolf, L.: Improved stereo matching with constant highway networks and reflective confidence learning. In: Proceedings of the IEEE conference on computer vision and pattern recognition. pp. 4641--4650 (2017)

\bibitem{shen2021cfnet}
Shen, Z., Dai, Y., Rao, Z.: Cfnet: Cascade and fused cost volume for robust stereo matching. In: Proceedings of the IEEE/CVF Conference on Computer Vision and Pattern Recognition. pp. 13906--13915 (2021)

\bibitem{shen2023digging}
Shen, Z., Song, X., Dai, Y., Zhou, D., Rao, Z., Zhang, L.: Digging into uncertainty-based pseudo-label for robust stereo matching. IEEE Transactions on Pattern Analysis and Machine Intelligence  (2023)

\bibitem{teed2020raft}
Teed, Z., Deng, J.: Raft: Recurrent all-pairs field transforms for optical flow. In: Computer Vision--ECCV 2020: 16th European Conference, Glasgow, UK, August 23--28, 2020, Proceedings, Part II 16. pp. 402--419. Springer (2020)

\bibitem{wang2021pvstereo}
Wang, H., Fan, R., Cai, P., Liu, M.: Pvstereo: Pyramid voting module for end-to-end self-supervised stereo matching. IEEE Robotics and Automation Letters  \textbf{6}(3),  4353--4360 (2021)

\bibitem{weinzaepfel2023croco}
Weinzaepfel, P., Lucas, T., Leroy, V., Cabon, Y., Arora, V., Br{\'e}gier, R., Csurka, G., Antsfeld, L., Chidlovskii, B., Revaud, J.: Croco v2: Improved cross-view completion pre-training for stereo matching and optical flow. In: Proceedings of the IEEE/CVF International Conference on Computer Vision. pp. 17969--17980 (2023)

\bibitem{xu2021bilateral}
Xu, B., Xu, Y., Yang, X., Jia, W., Guo, Y.: Bilateral grid learning for stereo matching networks. In: Proceedings of the IEEE/CVF Conference on Computer Vision and Pattern Recognition. pp. 12497--12506 (2021)

\bibitem{xu2022attention}
Xu, G., Cheng, J., Guo, P., Yang, X.: Attention concatenation volume for accurate and efficient stereo matching. In: Proceedings of the IEEE/CVF Conference on Computer Vision and Pattern Recognition. pp. 12981--12990 (2022)

\bibitem{xu2023iterative}
Xu, G., Wang, X., Ding, X., Yang, X.: Iterative geometry encoding volume for stereo matching. In: Proceedings of the IEEE/CVF Conference on Computer Vision and Pattern Recognition. pp. 21919--21928 (2023)

\bibitem{yang2018segstereo}
Yang, G., Zhao, H., Shi, J., Deng, Z., Jia, J.: Segstereo: Exploiting semantic information for disparity estimation. In: Proceedings of the European conference on computer vision (ECCV). pp. 636--651 (2018)

\bibitem{zhang2019ga}
Zhang, F., Prisacariu, V., Yang, R., Torr, P.H.: Ga-net: Guided aggregation net for end-to-end stereo matching. In: Proceedings of the IEEE/CVF Conference on Computer Vision and Pattern Recognition. pp. 185--194 (2019)

\bibitem{zhang2020domain}
Zhang, F., Qi, X., Yang, R., Prisacariu, V., Wah, B., Torr, P.: Domain-invariant stereo matching networks. In: Computer Vision--ECCV 2020: 16th European Conference, Glasgow, UK, August 23--28, 2020, Proceedings, Part II 16. pp. 420--439. Springer (2020)

\bibitem{zhang2020adaptive}
Zhang, Y., Chen, Y., Bai, X., Yu, S., Yu, K., Li, Z., Yang, K.: Adaptive unimodal cost volume filtering for deep stereo matching. In: Proceedings of the AAAI Conference on Artificial Intelligence. vol.~34, pp. 12926--12934 (2020)

\bibitem{zhao2023high}
Zhao, H., Zhou, H., Zhang, Y., Chen, J., Yang, Y., Zhao, Y.: High-frequency stereo matching network. In: Proceedings of the IEEE/CVF Conference on Computer Vision and Pattern Recognition. pp. 1327--1336 (2023)

\end{thebibliography}




\vspace{11pt}
\begin{IEEEbiographynophoto}{Weiqing Xiao} received the B.S. degree from 
  SHENYUAN Honors College of Beihang University in 2022. 
  He is currently pursuing the master's degree in School of Electronic 
  Information Engineering, Beihang University. His research interests include 
  computer vision and 3D vision.
\end{IEEEbiographynophoto}

\vfill

\end{document}